\documentclass[10pt, conference, compsocconf]{IEEEtran}
\usepackage{times}
\usepackage{epsfig}
\usepackage{graphicx}
\usepackage{amsmath}
\usepackage{amssymb}
\usepackage{multirow}
\usepackage{subfig}
\usepackage[pagebackref=true,breaklinks=true,letterpaper=true,colorlinks,bookmarks=false]{hyperref}
\begin{document}

\title{Bootstrapping Labelled Dataset Construction for Cow Tracking and Behavior Analysis}

\author{
\IEEEauthorblockN{Aram Ter-Sarkisov, Robert Ross, John Kelleher}
\IEEEauthorblockA{School of Computing,\\
Dublin Institute of Technology\\
Dublin, Republic of Ireland\\
453615@dit.ie}
}
\maketitle
\begin{abstract}
This paper introduces a new approach to the long-term tracking of an object in a challenging environment. The object is a cow and the environment is an enclosure in a cowshed. Some of the key challenges in this domain are a cluttered background, low contrast and high similarity between moving objects -- which greatly reduces the efficiency of most existing approaches, including those based on background subtraction. Our approach is split into object localization, instance segmentation, learning and tracking stages. Our solution is benchmarked against a range of semi-supervised object tracking algorithms and we show that the performance is strong and well suited to subsequent analysis. We present our solution as a first step towards broader tracking and behavior monitoring for cows in precision agriculture with the ultimate objective of early detection of lameness.
\end{abstract}

\IEEEpeerreviewmaketitle

\section{Introduction}
In the domain of modern animal husbandry the early detection and treatment of lameness is a serious and challenging problem. Lack of adequate treatment can lead to substantial losses for farms and reduced well-being for animals \cite{warnick2001effect, hernandez2001effect}. Detection and treatment of animal lameness has traditionally involved the hiring of expensive specialists after the disease has already become highly pronounced. Due to the negative consequences of late detection, there has recently been an increased interest in applying statistical and machine learning methods to lameness detection. These methods range from regressions and longitudinal studies (\cite{kamphuis2013applying, alban1996lameness}) through to neural networks and support vector machines \cite{pastell2007probabilistic, martiskainen2009cow, poursaberi2010real}. Analysis such as these mainly rely on 4 main sensor types: accelerometers, weight platforms, remote sensors and video cameras. In our work we are particularly focused on video data since it is convenient for both humans and cows;  there is no need for a lengthy installation of equipment, the equipment is cheap when viewed over the long term, and importantly for the animal the method is non-invasive. \\
\linebreak
The main drawback of video based analysis in this domain is the complexity in information retrieval: one needs to extract the animal's shape and behavior over a period of time and in relatively complex environments. As such, to the best of our knowledge, previous video recordings for lameness detection were performed in open space with good contrast between the background and the object, as in \cite{poursaberi2010real, song2008automatic} over a short period of time. In Section \ref{sec:previous} we discuss these and other results in greater detail.\\
\linebreak
It has been shown elsewhere that there are multiple features that correlate with lameness, i.e., gait, head tilt, weight distribution, and behavior. While each of these can in principle be analyzed through video, we suggest that analysis of behavior is particularly interesting, because difference in behavior between lame and so-called sound cows has been observed extensively (\cite{olechnowicz2011behaviour, alsaaod2012electronic,  kamphuis2013applying, cangar2008automatic}). For example it has been shown that the frequency and duration of actions like lying and walking correlates with lameness onset. \\
\linebreak
Our long term research goal is to monitor animal behavior directly from video to predict the early onset of lameness, which, given the above, is harder than with the use of accelerators or weight platforms from both scientific and technical points of view. Among other things, it cannot be approached in a straightforward manner (raw data $\to$ features $\to$ classification), because extraction of raw features and generation of a labelled corpus from video data full of challenges (poor lighting conditions, frequent occlusion, bad contrast) are problems that do not have a straightforward solution. In the current article we set out a method that combines deep learning algorithms, heuristic methods and an ensemble learning algorithm to track the movement of a cow and construct a labelled dataset.\\
\linebreak        
Our method is constructed and benchmarked against a new video corpus of animal behavior. We detail the content and construction of this video corpus in Section \ref{sec:data}. We then present the details of our novel approach to object tracking in Section \ref{sec:our_approach}. In Section \ref{sec:res} experimental results are presented together with a comparison to other relevant algorithms. Finally in Section \ref{sec:future} we outline our conclusions and plans for future work.    
\section{Previous Work}
\label{sec:previous}
The use of Computer Vision in monitoring animal behavior has a varied history. Rather than providing a comprehensive review of all such work here, we instead focus on the challenge at hand, i.e., the specific case of behavior tracking for bovines. 
To the best of our knowledge, there has to date been no direct overlap in the literature between object tracking and behavior-based cow lameness detection from video. However, there have been attempts to use machine learning algorithms like support vector machines (SVM) to directly predict cow lameness stage and lameness identification from video. In \cite{poursaberi2010real} cow's backposture was extracted from two sets ($n_1=28, n_2=66$) of cows shot on video while walking and scored by observers based on the presence of lameness features between 1 (no lameness) and 3 (severe lameness). 
The authors report a 96\% accuracy (percentage of correctly classified observations) of estimation using this method. Background subtraction in the publication was done in a relatively straightforward manner, because the video was shot outside at a convenient angle with little to no clutter in the background and very mild partial occlusion. Most importantly, contrast between the cow and the background is quite sharp, and there were no other cows in the video. \\
\linebreak
Earlier work \cite{song2008automatic} showed that cow's movement (locomotion) was examined in a manner similar to \cite{poursaberi2010real}. That work established high correlation (94.2\%) between hoof positions estimated by the camera and humans, and the difference in the position of the hooves between sound and lame cows. In addition to these two studies, there have been other publications on lameness detection based on video, but all of them were shot outside, with good contrast between background and object, making the task much easier then the problem we address in this paper.\\
\linebreak
Beyond the specific application to cow tracking, there has of course been significant research into object tracking. Of this work the kernel-based, semi-supervised and ensemble tracking algorithms have shown much promise. Tracking algorithms in these categories are some of the most popular, because they fulfil two important requirements: they do not require fully-labelled datasets for supervised learning and they generalize well to different problems. Semi-supervised algorithms include Tracking-Learning-Detection (TLD), developed by Kalal et al, \cite{kalal2012tracking}, a related Median Flow algorithm, also by Kalal et al, \cite{kalal2010forward}, Multiple Instance Learning  (MIL) (\cite{babenko2009visual, babenko2011robust}) with an extension built upon random forests, \cite{leistner2010miforests}. Kernel-based algorithms use a Kernel filter for similarity measure: \cite{comaniciu2003kernel}. One of the most recent approaches is Kernelized Correlation filter (KCF) by Henriques et al, \cite{henriques2012exploiting, henriques2015high}, that employs circulant matrices, several types of kernels (Gaussian, Linear, Ridge) and Fast Fourier Transform to learn a set of dense samples (all subwindows) from the tracked object. A big advantage of all these algorithms is that the user only needs to define the starting coordinates and the size of the bounding box(hence semi-supervised), therefore they are easy to test. Also recently, deep learning algorithms, such as Convolutional Neural Nets (ConvNN) were adapted for tracking tasks by Ma et al in \cite{ma2015hierarchical}. These ConvNNs use a popular VGG-Net-19 architecture (\cite{simonyan2014very}). A major disadvantage to these algorithms and the reason that we cannot apply them directly at this point is that they require a significant amount of labelled data. 
\section{Approach Overview}
\label{sec:overview}
In our work we have been interested in the automated bootstrapping of labelled dataset construction to cut down on the cost and technical challenges associated with building a large labelled dataset suitable to use with for example Deep Learning methods. Our approach to building this dataset is to build a predictive model that can track a particular cow through the video. We use a Random Forest classifier to do this tracking. Here we will give a high-level overview of the stages in the approach so as to help reader understand how the different components in our approach interact.\\ 
\linebreak
Our approach to building a tracking model for a given cow involves a number of steps. The first step is that we choose the cow in the video that is to be tracked. We do this by drawing a bounding box around that cow in the first frame of the video and label this bounding box as containing the target cow. We then also draw bounding boxes around all the other cows in the video and label these bounding boxes as distractor cows. We then sequentially process each frame in the video.\\ 
\linebreak
When processing each frame we first apply a model called CRFasRNN \cite{zheng2015conditional} to localize blobs of pixels that the model predicts as belonging to a cow or multiple cows. Because of this ambiguity, we then apply an edge detector called HED \cite{xie2015holistically} followed by a thresholding method (ISODATA \cite{ridler1978picture}) and this process isolates instances of cows within each blob. In other words, this process may segment a blob further into multiple cow instances. Note, that for now we do not address the problem of merging blobs but we will discuss this in future work.\\ 
\linebreak
Once we have extracted a set of cow instances from the frame, we then label each of these cow instances in the frame as belonging to either the target cow or one of the distractor cows. We do this by labelling each instance in the current frame with the label of the nearest instance in the previous frame. Using this process of frame analysis followed by the nearest neighbor instance labelling, we can track the target cow successfully through a short well-behaved video sequence. However, this approach doesn't scale to longer noisier videos. To do this we use the labelled short video to construct a training set for a random forest model that can track the cow through the longer more difficult video sequences. We build this training set for the random forest model by extracting 9 features from each instance in each frame. We then construct the training set by having one-row-per-instance-per-frame with each row labelled as being either the target instance or distractor instance. We then train and validate  the random forest model on this constructed dataset and then test this model on the portions of the video not used to construct the dataset. Having obtained the results, we manually identify true and false positives on a frame by frame basis. 
\section{Datasets}
\label{sec:data}
Our raw datasets consist of video data collected over a period of 14 days in a cowshed environment. Cameras observed enclosures which contained 10 individual animals. Cameras were mounted at a fixed angle to the animals, and in total 3TB of video data was collected. The data was collected and provided by the Irish National Agriculture and Food Development Authority (Teagasc), No labelling of the raw data was provided. \\ 
\linebreak
\begin{figure*}[t]
\begin{center}
\fbox{\includegraphics[width = \textwidth, height=150pt]{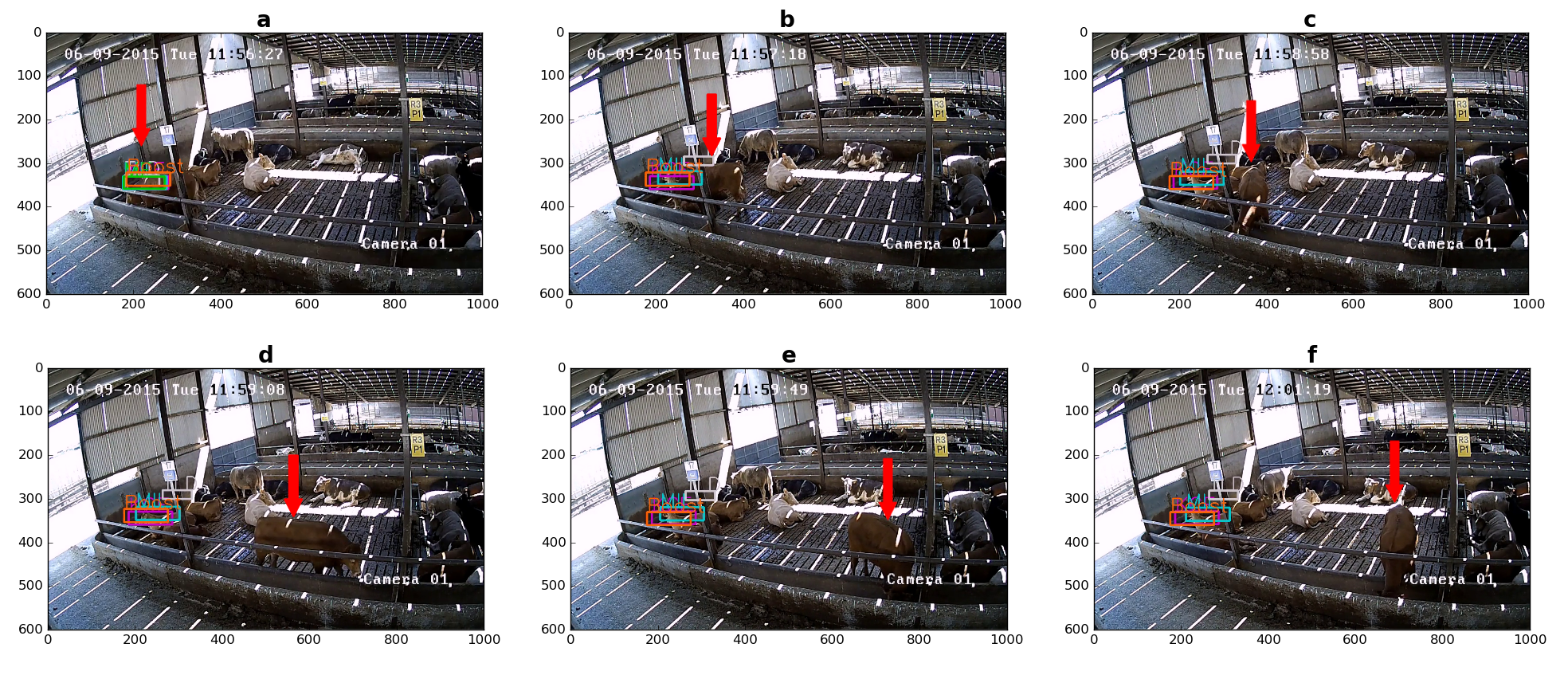}}
\end{center}
\caption{Performance of KCF (purple box), MIL (light blue box), TLD (white box), MedianFlow (green box) and Boosting (orange box) tracking algorithms on the test dataset \textbf{ch0106092015115543}. The red arrow points to the cow that these algorithms must track for the whole duration of the snippet. Each algorithm is initialized in the frame (a) with the features sampled from the user-defined bounding box that includes the greater part of the tracked cow. All six of them break down as the cow starts moving away from the starting position. The cow first ruminates close to the starting position in (b), then moves to the right of the vertical bar and stops to drink in (c), then continues its movement to the right in (d) in order to finally stop and start eating/drinking from the trough, in (e) and (f). As detailed in Table \ref{tab:snippets}, the total length of this snippet is 300 seconds. Best viewed in color.} 
\label{fig:other}
\end{figure*}
\linebreak
From this wealth of data we extracted a number of snippets, listed in Table \ref{tab:snippets}. As a video sequence is a time series, i.e. ordered data, each snippet is split into two consecutive subsets: training/validation and testing. Test datasets are usually a few times longer than training and validation datasets. Since cows move slowly, we only took every 10th frame from the video sequence. Therefore, in Table \ref{tab:snippets} we report both the number of frames and the length of the video. Names of datasets consist of the channel/camera id, date (day, month, year format) and the time in hour:minute:second format, with the actual time corresponding to the first second of the snippet. For example, \textbf{ch0106092015115543} means camera  number 1 shooting on June, 9, 2015 starting at 11:55:43.\\
\linebreak
Training and testing datasets were selected with the following objectives in mind: in the training set the cow that we want to track has to be fairly well visible for the whole duration (therefore they are quite short), so that the learning algorithm has enough correctly labelled features to train with. In the test set, on the other hand, most or at least some challenges should be present. While we recognize that these are simplifying assumptions in comparison to making a purely random selection of sequences from the data, we believe that this method is useful at this stage.\\ 
\linebreak
Given the nature of the problem and the recording environment, the dataset includes a number of challenges: 
\begin{enumerate}
\item Background: background in the video is generally dark and suffers from the low contrast, it is easy to confuse the background with animals, especially dark-skinned (black and brown), certain cows (especially black ones) are often not discernible even by a human eye,
\item Lighting: the lighting is generally low and uneven due to the presence of narrow and long gaps in the walls and ceilings. As a result, many cows have bright rectangular patches on their skin, often splitting the object in two or more parts. 
\item Objects: In every enclosure there are 10 cows of approximately the same size and different skin color, usually black, brown, white and striped (white and black), hence they are easy to confuse with each other. As cows are malleable objects, throughout the video their appearance changes substantially, from small while facing away from the camera and blending with the background to large and contrasting when standing perpendicular to the camera's direction. 
\item Occlusion: there are two types of occlusion in the dataset. First, the components of the cowshed and enclosure, like metal bars and concrete troughs that serve as boundaries of the enclosure. Secondly, due to the size of the enclosure, cows block each other from view much of the time, thus if a cow changes its action (walking$\to$standing, standing$\to$lying) while blocked away from the view, it is very challenging to identify this change automatically.
\end{enumerate}  
These features cause considerable trouble for existing tracking algorithms. This is exhibited in Fig. \ref{fig:other}, where we used five tracking algorithms implemented in OpenCV 3.1.0 library for Python 2 and mentioned in Section \ref{sec:previous}: MIL, TLD, KCF, Boosting and MeanFlow. Their drawbacks become evident after about first 70-100 frames (35-50 s) as the cow starts to move from its starting position. Trackers fail to account for the changing shape of the cow as it turns around and instead learn from other objects in the bounding box: a similar cow and the background. The second challenge are the metal bars (enclosure boundary) serving as a partial occlusion as the cow moves behind it. 
\section{Method}
\label{sec:our_approach}
\begin{figure*}[t]
\begin{center}
\fbox{\includegraphics[width = \textwidth, height = 130pt]{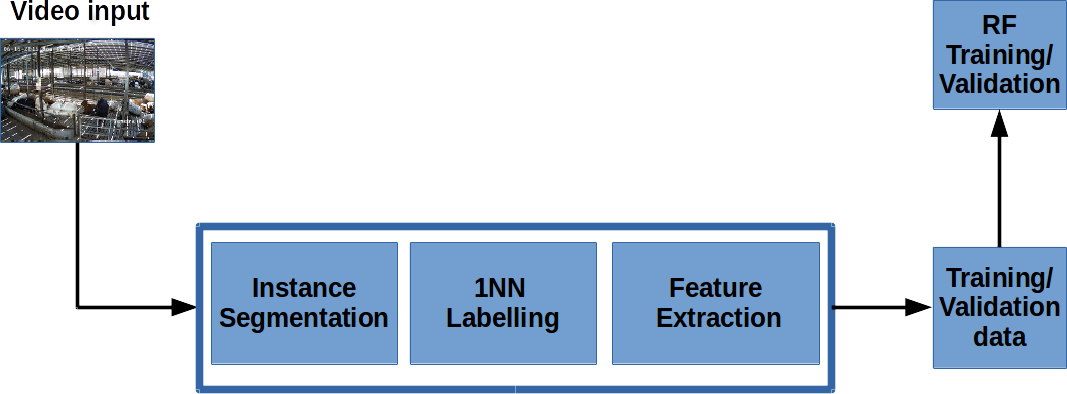}}
\end{center}
\caption{Flowchart of the first step of the algorithm. Video data is processed framewise: phases within the bounding box are repeated for the full duration of the video, adding labelled data points to the training dataset. Once the training dataset is fully built and manually cleansed of mislabelled observations, it is passed on to the Random forest (RF) algorithm for training and cross-validation. The final output of this stage is an RF classifier with optimal parameters for tracking a particular cow.}
\label{fig:train}
\end{figure*}

\begin{figure*}[t]
\begin{center}
\fbox{\includegraphics[width = \textwidth, height = 140pt]{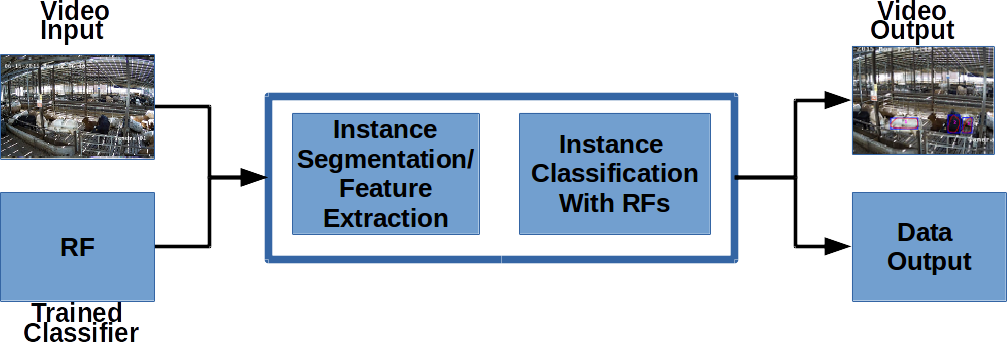}}
\end{center}
\caption{Flowchart of the second step of the algorithm. Video data is processed framewise: phases within the bounding box are repeated for the full duration of the video. RF classifier optimized in the first step processes objects segmented in the instance segmentation stage in each frame and labels them based on the features extracted from instances. Labels and instances are added to the video output. Some data is stored and output for further analysis.}
\label{fig:test}
\end{figure*}

An essential feature of many tracking algorithms is the dependence on the contrast between the tracked object and background, however cluttered it is. This poses a particular challenge for our project, because pixel intensity (hence the color) of large areas of the background, such as the floor, drinking troughs, metal bars and pathways between enclosures are very similar to that of many dark-skinned cows. Similarly, light-skinned cows are easily confused with patches of light passing through gaps in the wall and ceiling. Another problem is that cows of the same color are essentially similar, hence it is enormously challenging for a tracking algorithm to tell between two brown cows, especially if one of them blocks the other from camera view. The main idea of our approach for this reason is to extract contours of the tracked object instead of the background. As explained in Section \ref{sec:overview}, the algorithm consists of two steps: the first one does instance segmentation, feature extraction and training and validation, it is presented in Fig. \ref{fig:train} and the second does feature extraction and testing of a learning algorithm, and is presented in Fig. \ref{fig:test}. The instance segmentation phase is presented in Fig. \ref{fig:instance}.  
\begin{table*}
\centering
\caption{Size (in frames) and length (in seconds) of training/holdout and test sets.}
\begin{tabular}{|c|c|c|c|c|}
\hline
\multirow{2}{*}{\textbf{Title}}&\textbf{Size of training and} &\textbf{Size of test set}&\textbf{Length of training and}&\textbf{Length of}\\
&\textbf{holdout sets}&\textbf{and holdout sets}&\textbf{holdout sets}&\textbf{test set}\\
\hline
\hline
\textbf{ch0106092015115543}&107&600&43&300\\
\hline
\textbf{ch0406272015143027}&150&360&150&360\\
\hline
\textbf{ch0106292015090316}&92&750&91&765\\
\hline
\textbf{ch0710062015201033}&89&362&89&364\\
\hline
\textbf{ch0915062015120155}&47&600&25&605\\
\hline
\end{tabular}
\label{tab:snippets}
\end{table*}
\subsection{Framewise cow instance segmentation}
\begin{figure*}[t]
\begin{center}
\fbox{\includegraphics[width=\textwidth, height = 210pt]{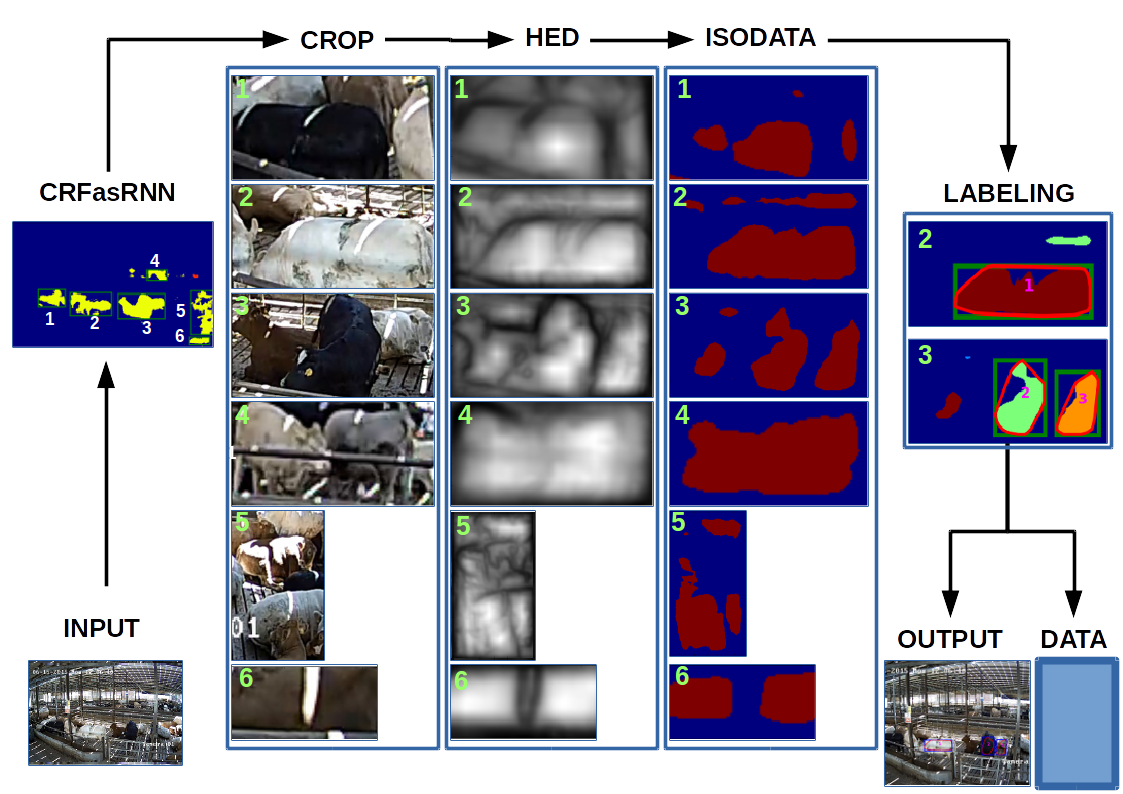}}
\end{center}
\caption{One full step of cow instance segmentation. The input is a frame 1000x600 pixels. The first step is localization with CRFasRNN. In the second step we crop all promising image insets that exceed a pre-set threshold value. In each inset we extract contours with HED. To obtain isolated objects from these contours we use an ISODATA thresholding function. Each object with size exceeding another pre-set threshold value is assigned an object status: we obtain its convex shape and a bounding box around it. We add these contours and bounding boxes to the input image and sample the data from the image, which is added to the dataset. The output is an image with overlaid bounding boxes and object contours.}
\label{fig:instance}
\end{figure*}

This is the first important phase in both steps of our approach. We start by localizing potentially interesting areas in the frame, and for this purpose we use a pre-trained deep learning algorithm: Conditional Random Fields as Recurrent Neural Network (CRFasRNN), recently introduced in a paper by Zheng et al in \cite{zheng2015conditional}. This is a combination of a fully convolutional network (FCN, see \cite{long2015fully}) and a conditional random field (CRF) with a layout of a recurrent neural network, RNN. This novel object segmentation algorithm returns a mask with the color of the pixels corresponding to the identified class, see Fig. \ref{fig:instance}. Yellow blobs in the mask correspond to the identified cow pixels (some cows in this video were misclassified as sheep and horses; to avoid further complications, we relabelled these pixels as cows and note that further training of the CRFasRNN is likely to increase accuracy). So far CRFasRNN has shown very high performance on our video data, compared to other algorithms, including FCN. ConvNN's architecture used for our approach is FCN-8, which is based on VGG-16 (\cite{simonyan2014very}). This identification of elements in the raw image is by far the most computationally intensive part of our model, taking $\sim$50s/frame on the CPU and $\sim$7s/frame on GPU (Tesla K40). \\
\linebreak
Once we have localized the blobs with potentially interesting objects, we get a bounding box around those larger than an optimal threshold empirically found to be 800 pixels and extract this patch from the original image. What we want now is to identify instances of cows, since CRFasRNN does not report the number of instances, but only that the  observations in these blobs are consistent with the class `cow'. As explained above, background subtraction is intractable in cowsheds due to very noisy and low-contrast background. Instead, we attempt to extract contours of objects. This comes from the observation that, even if the objects have a similar pixel intensity, the value of the derivative at the boundary has the potential for delineating between them.\\ 
\linebreak
For contour detection we found that the Holistically-Nested Edge Detector (HED), recently introduced by Xie and Tu in \cite{xie2015holistically} performs strongly by filtering out most of the noise in the image. Therefore in the next stage we run each localized area in the image through HED to obtain edges. The output of HED is an image with the darker pixels corresponding to more important edges. Since we only need the most important of them, we want to split each contour image into two disjoint subsets: objects and background. We do this by using the threshold detection algorithm ISODATA, introduced by Ridler and Calvard in \cite{ridler1978picture}. 
A combination of these two methods produces a set of isolated objects that we label as such and add to the training set should their area exceed 800 pixels (also found empirically). We also get the objects' convex hulls (to track the change in the cows' shape) and bounding boxes.This step is much less computationally expensive, taking 5-7s on a CPU or $<1$s on a GPU (Tesla K40).\\  
\linebreak
Our approach takes CRFasRNN's output a step further by offering a solution for instance segmentation: blobs from CRFasRNN's output merely tell us that there are cows in this locality. We extract the hypothesis for the number of cows and the approximation for their shapes and locations. In Fig. \ref{fig:instance} we present the flowchart of the instance segmentation step.             
\subsection{Feature Extraction}
The process detailed in the previous subsection is applied to every frame in the video sequence. We combine that process with a simple 1-Nearest Neighbor (1NN) tracking method to automate feature extraction from instances identified in the first phase of processing. In the very first frame of the sequence, we select the cow we want to track and assign it label 1; all other objects are labelled with 0. For all other frames, once we have all the instances, we get a simple distance matrix, where the number of rows is equal to the number of instances in the current frame, and the number of columns to the number of labelled instances in the previous frame, hence each entry in the matrix is a distance from every instance in this frame to every labelled instance in the previous frame. Once we have this matrix, we assign every instance in this frame the label of the nearest instance from the previous frame. This heuristic approach is  too simple for any serious tracking problem, and therefore our training databases are very small (50-100 frames) and well-behaved, i.e. the cow we want to track is well visible throughout the video of the training set. Once we have labelled all instances in the frame, we extract features from them, which are added to the training dataset in the correct order (i.e. concatenated with the previous data). In total, we use 9 features from three types: 
\begin{enumerate}
\item Pixel intensity: We use the instance's centroid as the mean and $\sigma^2=5$ to sample 100 5x5 patches in each object; after averaging over their pixel intensities we get a vector of features: overall mean, maximum and three quartiles.  
\item Size: we use the size of the bounding box around the object. This is motivated by the fact that in the previous stage we ignored small objects (under 800 pixels), and that our instance segmentation algorithm tends to find large portions of cows.
\item Location: we store the $(x,y)$ coordinates of the centroid of the bounding box as the distance feature. During the training and test phases, we find the distance between every instance's centroid and the centroid of the previous observation of the tracked cow. This coordinate difference is the actual feature used.        
\end{enumerate}       
Finally, once we have collected all the data from the training video sequence, we manually clean up the training dataset by removing mislabelled (false positive) observations and pass the correctly labelled dataset to the training algorithm.\\
\linebreak
\subsection{Training}
With features extracted for each cow, we trained a classifier to automatically identify an individual animal in a video frame. We selected the Random Forest (RF) classifier, attributed to Breiman \cite{breiman2001random} to learn the features. Originally in each training set the proportion of positive (tracked cow) observations is about 16 \%. To provide the classifier with more data, we sampled out about 50 \% of negative (distractor cows) observations, thus increasing the positive data points to about a third of the training database. As the data is essentially a time series, we train the algorithm on the first $K$ observations and validate on the remaining $n-K$. The second contribution of this paper, after instance segmentation algorithm, is optimizing a classifier based on the training, validation and testing output. We found that RF with 300 trees, cross-entropy error function, using all features during training, with bootstrap samples and out-of-bag samples for generalization do the best job on our data. Training of a single forest takes a very small amount of time, $\sim$ 5 seconds.
\begin{table*}
\centering
\caption{Performance of the five algorithms available in the OpenCV library on the test datasets compared to ours. Values in each column are the percentage of frames the algorithm identified the tracked cow, i.e. recall. Bold are the best algorithm for this test dataset.}
\begin{tabular}{|c|c|c|c|c|c|c|}
\hline
\textbf{Video}&\textbf{KCF}&\textbf{TLD}&\textbf{MIL}&\textbf{Boost}&\textbf{MF}&\textbf{Ours}\\
\hline
\hline
\textbf{ch0106092015115543}&0.21/0.2&0/0&0.17/0.2&0.2&0&\textbf{0.98/0.54}\\
\hline
\textbf{ch0406272015143027}&\textbf{1}/0.76&0.56/0.41&\textbf{1/0.88}&0.93/0.17&0.4/0.41&\textbf{1}/0.77\\
\hline
\textbf{ch0106292015090316}&\textbf{1/1}&0.02/0&\textbf{1}/0.56&1/0.35&0.5/0.13&0.7/0.47\\
\hline
\textbf{ch0710062015201033}&\textbf{1/1}&0.44/0.16&\textbf{1/1}&0.78/\textbf{1}&0.4/0.16&0.83/0.08\\
\hline
\textbf{ch0915062015120155}&0.36/0.34&0.25/0.4&0.36/0.34&0.36/0.34&0.28/0.33&\textbf{0.56/0.85}\\
\hline
\end{tabular}
\label{tab:tracking_other}
\end{table*}
\begin{table*}
\centering
\caption{Results on the test sets obtained using Random Forest with the best parameters trained on datasets in Table \ref{tab:snippets}. TP: true positives, FP:false positives, TN:true negatives, FN: false negatives, P:precision, R:recall. In the last column main challenges that the algorithm faced in the video are listed.}
\begin{tabular}{|c|c|c|c|c|c|c|c|}
\hline
\textbf{Video}&\textbf{TP}&\textbf{FP}&\textbf{TN}&\textbf{FN}&\textbf{P}&\textbf{R}&\textbf{Challenges}\\
\hline
\hline
ch0106092015115543&328&6&4928&272&98.2\%&54.6\%&Partial occlusion, dim and uneven lighting, low contrast\\
\hline
ch0406272015143027&275&0&3119&85&100\%&76.8\%&Cluttered background\\
\hline
ch0106292015090316&355&131&9529&395&70\%&47\%&Partial Occlusion, very low contrast, dim lighting\\
\hline
ch0710062015201033&30&6&2945&330&83\%&8\%&Partial occlusion, very uneven lighting,low contrast\\ 
\hline
ch0915062015120155&419&300&6048&71&56\%&85\%&Partial and full occlusion, very uneven lighting, cluttered background\\
\hline
\end{tabular}
\label{tab:results_tracking}
\end{table*}   	       
\section{Results}
\label{sec:res}
Results from testing the RF classifier, precision and recall, are summarized in Tables \ref{tab:tracking_other} and \ref{tab:results_tracking}. Test sets, which are taken from the same video, are different to the training set in a number of ways: there are many issues that an algorithm must handle, such as full and partial occlusion, bad lighting, low contrast, cluttered background and other. For comparison we use the five trackers mentioned above: TLD, MIL, KCF, MedianFlow and Boosting on each test set. In three datasets our approach achieves the highest precision and in two - the highest recall rate. Its strength is particularly well visible on \textbf{ch0106092015115543} and \textbf{ch0915062015120155}, where the tracked cow moves around. Other either immediately loose it (as TLD in \textbf{ch0106092015115543}) or confuse it with the background as soon as the cow leaves the area where the tracking started. We consider this to be a specific strength of our algorithm. On two sets where our approach underperformed (e.g.\textbf{ch0710062015201033}) the problem is related to the generalization capacity of the classifier: the cow does not move much, but its features are too easily confused with those of other objects.
\section{Discussion and Future Work}
\label{sec:future}
In this article we have presented a new tracking algorithm developed for tracking malleable objects (cows) in a challenging environment (enclosures in a cowshed). The ultimate goal of this project is to identify lameness in cows at an early stage; successful cow tracking is the first stage in this project. This article has three main contributions: 
\begin{enumerate}
\item Framewise instance segmentation,
\item Optimal Random Forest algorithm,
\item Construction of a large dataset for further analysis of cows' behavior
\end{enumerate}
From here there are three main directions in which we would like to take the development of this algorithm: 
\begin{enumerate}
\item Improvement of instance segmentation step. Currently we use a combination of two deep learning, thresholding and a labeling algorithm. Although they do the job reasonably well, there is enough space for improvement. 
\item Improvement of generalization. Although Random Forest does a good job with the tracking, it does not always confidently generalize to any angle at which the cow faces the camera. We therefore intend to retrain algorithm like CRFasRNN to get it to track objects (transfer learning) using the segmentation information from the previous step. 
\item Construction of cow behavior dataset. In addition to tracking the cows' movement, we need to track their behavior, as it correlates with the physiological condition (overall, lame cows lie for longer periods of time). For this purpose we will be training a large deep learning algorithm like LSTM (\cite{hochreiter1997long}) or a similar recurrent neural network. 
\end{enumerate}
\label{sec:future}
{\small
\bibliographystyle{IEEEtran}
\bibliography{IEEEabrv,article_tracking}
}
\end{document}